\documentclass[letterpaper, 10 pt, conference]{ieeeconf}  % Comment this line out if you need a4paper

\IEEEoverridecommandlockouts                              % This command is only needed if 
\usepackage{graphicx}
\usepackage{textgreek}
\usepackage{amsmath}
\usepackage[font=footnotesize]{caption}
\usepackage{subcaption}
\usepackage{booktabs}
\usepackage{multirow}
\usepackage{array}
\usepackage{mathtools}
\usepackage{comment}
\usepackage{amsmath}
\usepackage{amsfonts}
\usepackage{amssymb}
\usepackage{xcolor}
\usepackage{hyperref}
\usepackage{cite}
\usepackage[linesnumbered,ruled,vlined]{algorithm2e}

\SetCommentSty{mycommfont}

\SetKwInput{KwInput}{Input}                % Set the Input
\SetKwInput{KwOutput}{Output}              % set the Output

\title{\LARGE \bf
GHACPP: Genetic-based Human-Aware Coverage Path Planning Algorithm for Autonomous Disinfection Robot
}

\author{Stepan Perminov, Ivan Kalinov and Dzmitry Tsetserukou% <-this % stops a space
\thanks{All authors are with the Intelligent Space Robotics Laboratory, Space CREI, Skolkovo Institute of Science and Technology, Moscow, Russian Federation.
{\tt \{stepan.perminov, d.tsetserukou\}@skoltech.ru, ivan.kalinov@skolkovotech.ru}}%
}

\begin{document}
\maketitle
\thispagestyle{empty}
\pagestyle{empty}

%%%%%%%%%%%%%%%%%%%%%%%%%%%%%%%%%%%%%%%%%%%%%%%%%%%%%%%%%%%%%%%%%%%%%%%%%%%%%%%%
\begin{abstract}
Numerous mobile robots with mounted Ultraviolet-C (UV-C) lamps were developed recently, yet they cannot work in the same space as humans without irradiating them by UV-C. 
This paper proposes a novel modular and scalable Human-Aware Genetic-based Coverage Path Planning algorithm (GHACPP), that aims to solve the problem of disinfecting of unknown environments by UV-C irradiation and preventing human eyes and skin from being harmed. 

The proposed genetic-based algorithm alternates between the stages of exploring a new area, generating parts of the resulting disinfection trajectory, called mini-trajectories, and updating the current state around the robot.
The system performance in effectiveness and human safety is validated and compared with one of the latest state-of-the-art online coverage path planning algorithms called SimExCoverage-STC.
The experimental results confirmed both the high level of safety for humans and the efficiency of the developed algorithm in terms of decrease of path length (by 37.1\%), number (39.5\%) and size (35.2\%) of turns, and time (7.6\%) to complete the disinfection task, with a small loss in the percentage of area covered (0.6\%), in comparison with the state-of-the-art approach.
\end{abstract}

%%%%%%%%%%%%%%%%%%%%%%%%%%%%%%%%%%%%%%%%%%%%%%%%%%%%%%%%%%%%%%%%%%%%%%%%%%%%%%%%

\section{Introduction}

\subsection{Motivation}

In the face of the COVID-19 world-girdling pandemic, it has become apparent how important the disinfection of premises is to our lives. Among existing ways of sanitizing premises, using UltraViolet-C (UV-C) lamps is the most effective in terms of killing not only SARS-CoV-2 but also many other harmful bacteria and viruses. 
Whereas statically irradiating UV-C lamps cannot disinfect the target area uniformly \cite{tiseni2021uv}, many mobile service robots with mounted UV-C lamps have been developed recently \cite{chanprakon2019ultra, tiseni2021uv, UVDrobots}. All of them have a completely open lamp arrangement, which limits the scope of their use to unmanned areas and requires control of the disinfection process, since ultraviolet irradiation is dangerous to human eyes and skin.

To overcome this restriction, we developed an autonomous indoor mobile disinfection robot, called UltraBot, with two separate blocks of UV-C lamps with non-circular disinfection shape \cite{ultrabot} (more detailed description is presented in System Overview). In this manner, to conduct disinfection efficiently and safely for humans, an approach to the UltraBot path planning must be developed in accordance with the shape of the disinfection zone and the configuration of the lamps.

%To overcome this restriction, we developed an autonomous indoor mobile disinfection robot, called UltraBot, with two separate blocks of UV-C lamps \cite{ultrabot}. Its UV-C lamp arrangement allows performing disinfection of side areas without irradiating an area in front of and behind the robot. On the other hand, this kind of UV-C lamp arrangement provides a nonstandard, non-circular disinfection shape.
% Thus, to conduct disinfection both efficiently and safely for humans, it is necessary to develop an approach to UltraBot path planning to make the robot move along optimal trajectories, satisfying safety and efficiency requirements.
%In this manner, to conduct disinfection efficiently and safely for humans, an approach to the UltraBot path planning must be developed to make the robot move along trajectories that satisfy safety and efficiency requirements.

\begin{figure}[!t]
\vspace{-0.5em}
\centering
\begin{subfigure}{0.9\linewidth}
\includegraphics[width=\textwidth]{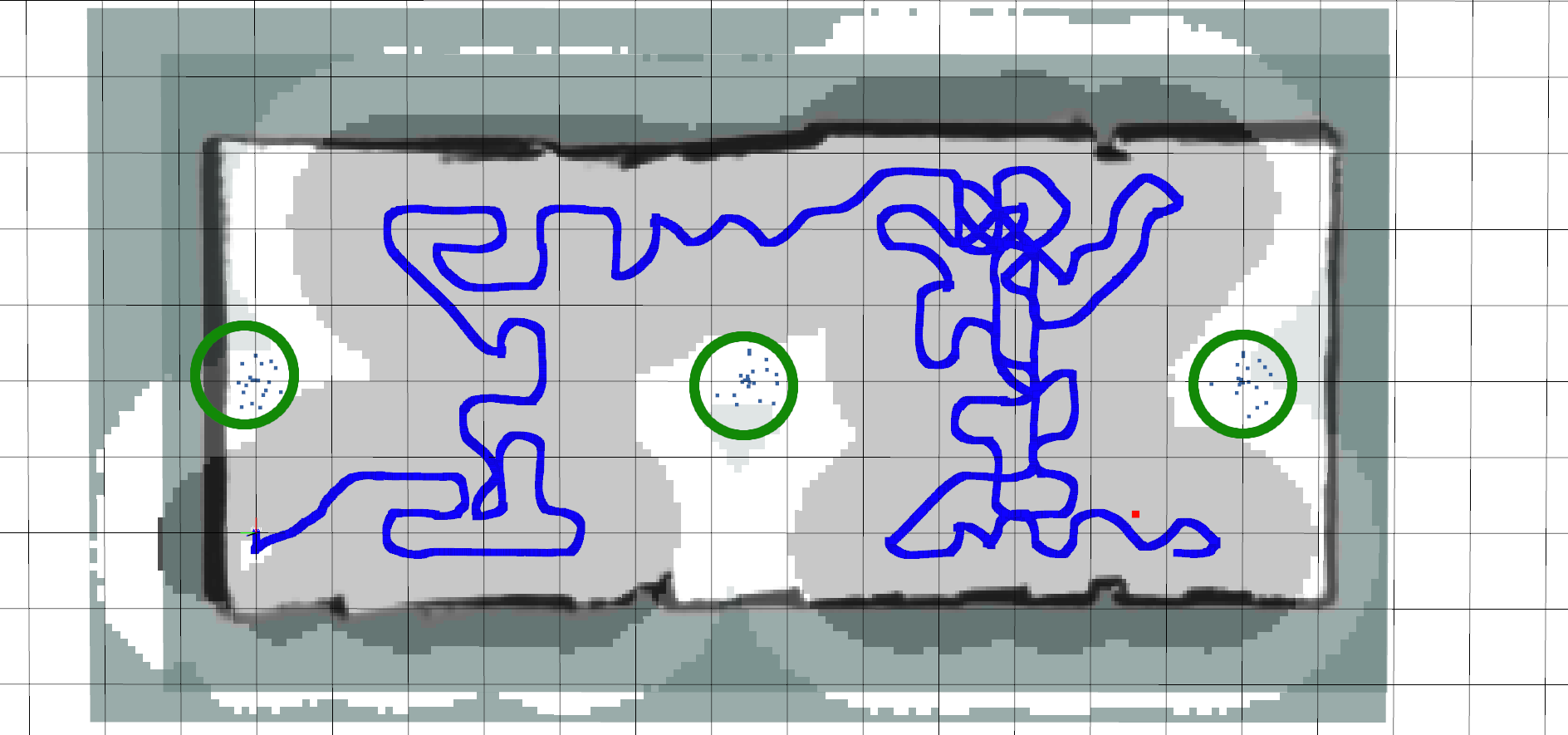}
\caption{GHACPP: Area coverage = 80.3\%, Path length = 86.95 m.}
\end{subfigure}
\begin{subfigure}{0.9\linewidth}
\includegraphics[width=\textwidth]{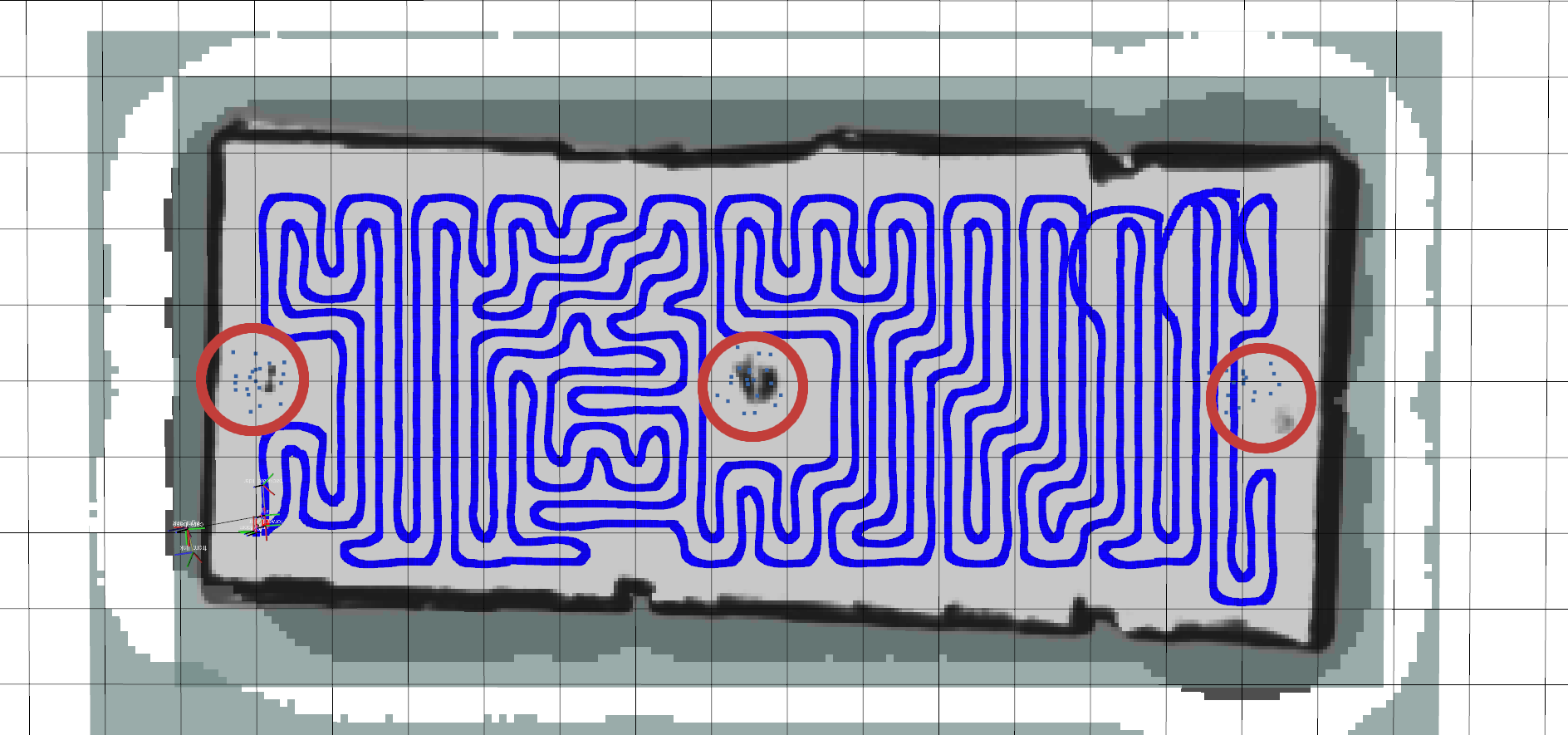}
\caption{SimExCoverage-STC: Area coverage = 100.0\%, Path length = 232.1 m.}
\end{subfigure}
\caption{Algorithm trajectories (blue lines) in the empty area. The dark shaded area is the disinfected area. The irradiation-free area is marked in white. Irradiated people in protective clothing are circled in red, unirradiated people are circled in green.}
\label{fig:exp2_trajectories_human}
\vspace{-2em}
\end{figure}

\subsection{Problem statement}
\label{problem}

Nowadays, there are many types of effective path planning algorithms. Luo et al. presented an improved ant colony algorithm \cite{luo2020research}, which showed better performance even in comparison with the method based on ant colony optimization with artificial potential field \cite{liu2015robot}. Neural RRT* \cite{wang2020neural} is based on the Convolutional Neural Network (CNN) model and the Rapidly-exploring Random Tree Star (RRT*) algorithm, which allows %achieving nonuniform sampling in the path planning process and
reducing the computational time and memory used for the path generation. Wang et al. proposed a novel non-uniform sampling technique \cite{wang2020optimal} to solve the problem of optimal path planning by using Generalized Voronoi Graph and Multiple Potential Functions.

However, these state-of-the-art algorithms are aimed at solving the problem of finding the shortest or optimal path in a space with obstacles only. They could not provide high-quality disinfection of premises by the robot with nonstandard, non-circular disinfection shape. In our case, the desired path must simultaneously meet the following set of requirements:

\begin{enumerate}
    \item Absence of collision with obstacles.
    \item Minimal number of repeatedly visited areas and repeatedly disinfected areas.
    \item Ability to consider the complexity of the robot's disinfecting zone shape for effective disinfection.
    \item Minimal number of turns to safe battery power and to reduce the time of the whole disinfection task.
    \item Avoidance of people irradiation by UV-C lamps.
\end{enumerate}

\subsection{Related works}

\subsubsection{Coverage path planning}
The robot path planning task for effective premises disinfection refers to Coverage Path Planning (CPP). For the CPP task, a mobile robot should cover the area in the most efficient possible way, saving the robot's energy and time. Many everyday applications of mobile robots require coverage path planning algorithms, such as autonomous vacuum cleaning, window cleaning, etc.  %Besides usual tasks, the CPP is relevant for painter robots \cite{hassan2019ppcpp}, UAV \cite{theile2020uav}, and reconfigurable robots \cite{le2020evolutionary}.

One of the first grid-based algorithms was the algorithm, called Spanning Tree Covering (STC) \cite{gabriely2001spanning}, which subdivides the work-area into disjoint cells corresponding to a square-shaped tool. %, and then follows a spanning tree of the graph induced by the cells while covering every point precisely once.
However, this algorithm leaves a lot of uncovered areas near the obstacle borders. %Improved version of the algorithm \cite{gabriely2003competitive} allows to visit partially occupied disjoint cells too, but still leaves many uncovered regions. A Spanning Tree-based Competitive and Truly Complete Coverage Algorithm (ST-CTC) \cite{guruprasad2015st} allows the robot to follow the obstacle border, thereby increasing the covered space percentage.
The Backtracking Spiral Algorithm \cite{gonzalez2005bsa} is a
coverage strategy for mobile robots based on the use of spiral filling paths. This approach provides high performance in cases of a simple environment. %, where the size of obstacles is a multiple of the size of the robot.
Otherwise, the robot makes a lot of overlaps to achieve the desired covering space percentage.

Cellular decomposition methods split the target area into regions which then are covered by a simple boustrophedon motion \cite{galceran2013survey}. %These algorithms work mostly in known and only in static environments.
Performance of some of the mentioned algorithms was compared in \cite{bormann2018indoor}. Cai et al. presented a method that made a mobile robot perform a simple boustrophedon or spiral motion until it faces the already cleaned area or an obstacle \cite{cai2014research}. This algorithm works in fully known environments only.
The algorithm \cite{song2018varepsilon} is built upon the concept of an Exploratory Turing Machine (ETM), which acts as a supervisor to the autonomous vehicle to guide it with adaptive navigation commands. %This algorithm works in both known and unknown environments.

In works \cite{yakoubi2016path}, \cite{prayash2019designing} a genetic algorithm (GA) was applied to solve the coverage path planning problem. However, the target covered area was discretized by discs as a shape of the robot. %instead of rectangles or squares to facilitate the robot's motion.
This area representation does not reflect the real space dimensions, which leads to a low covered space percentage. %The proposed algorithm works efficiently only if the robot starts from the environment corners.
Despite all its disadvantages, the genetic algorithm allows working in unknown environments and considering many conditions while planning paths, including the conditions into the cost function. %The genetic algorithm has great potential when it is equally important to cover the entire area while also quality disinfection and not exposing people to UV-C lamps. 

\subsubsection{Human-aware path planning}

Since the disinfection robot must perform path planning and follow it in such a way that it does not expose people to its UV-C lamps, human-aware path planning techniques should be highlighted.

%Some human-aware path planning algorithms are focused on the robots behaving in the most socially acceptable way possible. 
Vega et al. presented the algorithm, that determined if the space affordances, created by including certain objects with which humans often interact, were being used as activity spaces to consider them forbidden for navigation \cite{vega2019socially}. %Moreover, the algorithm clusters humans into groups according to their social interactions. 
Despite its advantages, the described algorithm doesn't work in crowded, dynamic environments, as it was developed to work in a static environment only.

Kuanqi Cai et al. proposes a human-aware motion planning algorithm \cite{cai2019adaptive} based on the adaptive sampling method to avoid the robot going into crowded areas and improve robot acceptance in the crowded public environment. %The algorithm predicts human movements using their current velocity values and directions using Risk-RRT*.
However, the algorithm doesn’t consider interpersonal relationships to perform prediction.

Perez et al. proposed an approach to learn path planning for robot social navigation by \cite{perez2018learning}. The problem was formulated as a classification task where a Fully Convolutional Network (FCN) is used to learn to plan a path to the goal in the local area of the robot in a supervised way. %Combined with an RRT* planner, the approach aims to solve the task and ensure collision-free paths efficiently.
However, the proposed approach cannot work in dynamic environments efficiently due to the absence of human trajectories' prediction.
 
Fang et al. proposed a global path planning method based on the global scope of pedestrian perception and multi-layer cost-maps \cite{fang2020human}.
%Multi-layer dynamic cost maps are generated containing the social cost at different time-steps, based on pedestrian trajectory prediction, which provides social constraints for global path planning.
Despite all its advantages, the target algorithm fails if the goal cannot be reached due to human appearance. 

The planning algorithm for the disinfection robot should make the robot disinfect as much space as possible, moving far away from people until humans get out of the way.

\subsection{Contribution} 

In order to overcome all the disadvantages of the previously developed algorithms, we propose a novel online Genetic-based Human-Aware Coverage Path Planning (GHACPP) algorithm for the disinfection robot with a specific UV-C lamp arrangement that disinfects premises by UV-C irradiation safely for humans. The word online means that the algorithm must constantly solve the Path Planning Coverage Region (PPCR) problem by generating motion trajectories as information about the space around the robot to be disinfected is updated. Our algorithm allows getting highly efficient robot trajectories along with considering a set of conditions and restrictions formulated in \autoref{problem} using the penalty system in our cost function.

Performance of the proposed algorithm was validated by comparison with one of the latest state-of-the-art online coverage path planning algorithms called SimExCoverage-STC \cite{falaki2020simultaneous} by solving an online PPCR problem for an autonomous mobile robot. The performance was evaluated based on the minimized function of parameters that represent the path planner effectiveness: 
\begin{enumerate}
    \item Area coverage, \%: The percentage of the disinfected area from the area, defined by the robot during its disinfection mission in previously unknown space.
    \item Path length, m: The global path length obtained after a set of several map extension - area exploration steps.
    \item Number of turns: A number of rotations the robot makes during the disinfection process.
    \item Total turn angle, rad: The total angle by which the robot turns while following the generated path.
    \item Time, sec: Time spent on performing the whole disinfection task, including both path planning and path following actions.
\end{enumerate}

\section{System Overview}

\begin{figure}[!t]
\vspace{0.5em}
\centering
%\vspace{-2.5em}
% \includegraphics[scale=0.45, width=0.33\textwidth]{figures/HQ Render Text.png}\label{true}
\includegraphics[width=0.35\textwidth]{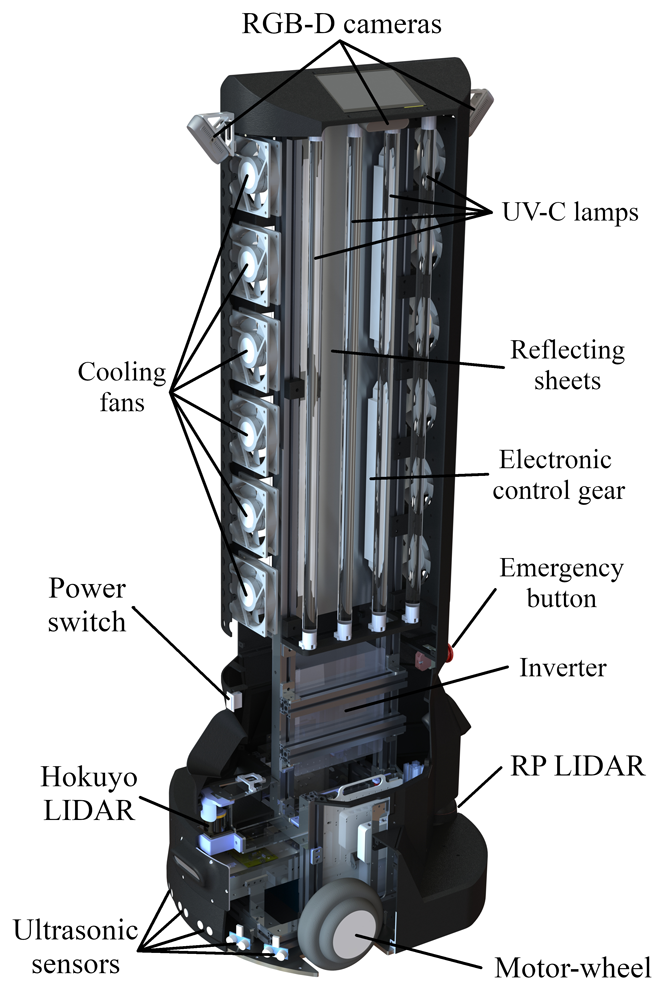}\label{true}
\caption{Decomposition of the UltraBot robot structure.}
\vspace{-1.5em}
\label{fig:isometric}
\end{figure}

%  In contrast to typical UV-C lamp placement in a cylindrical shape, we split the lamps into two opposite sides of the robot to restrict the UV-C emitting area to only 180 deg. Overall, it allows the robot to operate concurrently side by side with humans without downtime.  Eight lamps are installed on two opposite sides (4 lamps in the stack). Phillips TUV series lamps were chosen. Each lamp has 100 $\mu W/cm^2$ of UV-C irradiation at a distance of 1 meter and 30W electrical power specifications. The control is carried out using relays that commute AC voltage to  Electronic Control Gear (ECG) drivers.  An array of 12 DC fans supplemented the robot for additional air disinfection.
As a light source for disinfection, UV-C mercury vapor lamps of Phillips TUV series  were used (Fig. \ref{fig:isometric}). The UV-C fluorescent lamps are working at 253.7 nm wavelength, which has proven the highest germicidal effectiveness \cite{buonanno2020far}. Eight lamps are installed on two opposite sides (4 lamps in the stack) of the robot, and each stack can be used separately. Each illuminant has 100 \textmu{W/cm$^2$} of UV-C irradiation at a distance of 1 meter and 30 W of electrical power. Therefore, each side has 400 \textmu{W/cm$^2$} of irradiation, which is enough to kill 90\% of microorganisms \cite{bv2006ultraviolet}. % The control is carried out using relays that commute the alternating current (AC) voltage to Electronic Control Gear (ECG) drivers.
%and 120 W of power consumption

\subsection{Hardware}

%Strong attention was placed on the safety of neighboring humans. In contrast to typical UV-C lamp placement in a cylindrical shape, the lamps were split into two opposite sides of the robot, thus, the UV-C emitting area would be restricted to only 180$^\circ$. Overall, it allows the robot to operate concurrently side by side with humans without downtime.  As a result of the lamp stack placement facing each other, the lamps evidently will not work at full power because of the angle of glowing. The lamp module is supplied with a reflective shield made of celled anodized aluminum to avoid the loss of disinfection power \cite{ultrabot}.  

%In order to disinfect the air and add the ``safe mode" of work, i.e., when the light-blocking shield fully covers the UV-C bulbs in the future version, the robot was supplemented by an array of 12 DC fans. The fans are located lengthwise of the lamps, providing the significant airflow volume to pass through the maximum possible illuminating area. They are intended to work as the UV sterilizer in the situations when the robot is passing the humans. Thus, there will not be a loss in productiveness. During this work mode, the robot blows air stream through the disinfection system, functioning alike with a quartz lamp, simply maintaining airflow with the ionized particles and disinfecting the air.

Sensors have been chosen to produce a high quality of the object detection and sensing area. The Hokuyo Light Detection and Ranging (LiDAR) sensor aims implementation of simultaneous localization and mapping featuring high precision and update frequency. It is located in front of the robot's body. The RP2 LIDAR sensor determines rear collision to achieve the robot's 360 deg. field of view. Ten ultrasonic sensors and four Intel RealSense RGB-D cameras are used to detect collisions and people, respectively, to safely avoid them. % RBG-D cameras provide two types of data, such as RGB (red, green, blue), which identifies color images, and D (depth), which represents depth information.
% collision detection and obstacle avoidance.
Ultrasonic sensors are located beneath LIDARs to detect low obstacles such as soles of feet and doorsills, and transparent objects. RGB-D cameras are applied to detect the person approaching the upper part of the robot. Human detection is necessary for the emergency shutdown of lamps if someone is near the working robot and for the robot path planning while performing human-safe disinfection. %The communication protocol of high and low-level controllers is implemented with a USB virtual COM interface.
The robot's control system is based on the high-level controller (Intel NUC computer with Core-i7 processor) and low-level controller board (STM32). % for ultrasonic sensors, UV-C lamps control, LED control, and battery statistics. 

%The low-level controller board provides interfaces for ultrasonic sensors, UV-C lamps control, LED control, and battery statistics \cite{ultrabot}. 

\subsection{Shape of disinfection zone}

In order to find an effective disinfection zone, we conducted a series of real-word  experiments for detection of effective disinfection range and count Total Bacteria Count (TBC) depending on distance from the robot \cite{ultrabot_zone}. We received that TBC was equal to 100\% for distance of 0.3 meter, 86,9\% for distance of 0.6 meter, 84,6\% for distance of 0.9 meter. We chose the value at a distance of 0.9 meters as the main for disinfection due to it was a high value for distant disinfection according to Philips data sheet \cite{bv2006ultraviolet}.

\begin{figure}[h]
\vspace{-1em}
\centering
\includegraphics[width=0.45\textwidth]{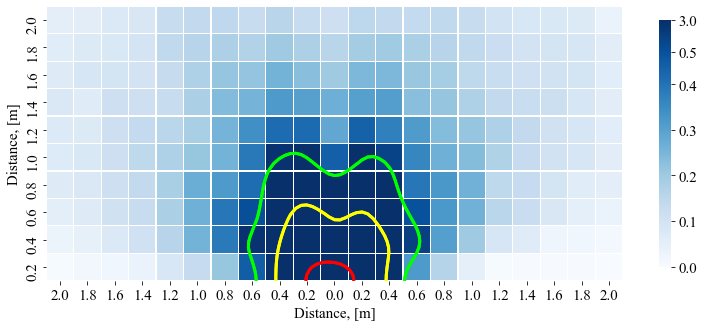}\label{true}
\vspace{-0.5em}
\caption{Effective disinfection zones with outer borders - 100\% (red), 86.9\% (yellow) and 84.6\% (green). Top view.}
\vspace{-1em}
\label{fig:luminocity_heatmap_plus_contour_plus_zones}
\end{figure}

In order to determine the robot's effective disinfection zone shape, we installed 4 fluorescent lamps (30W) into the robot in the same positions as UV-C ones to investigate luminosity distribution depending on the location relative to the lamps from one side of the robot. A normalized and then summarized luminosity value equals 0.54 that corresponds to 0.9 m distance from the robot fluorescent lamps is taken as a threshold value to reconstruct the outer border contour of the effective disinfection zone, provided in Fig. \ref{fig:luminocity_heatmap_plus_contour_plus_zones} by the green line. 
According to the lamp efficiency level, information on the distribution of disinfection zones will allow the robot to manage the disinfection process most effectively and achieve the desired disinfection level of the target premises. For this purpose, we also assigned the threshold luminosity values for different disinfection performances from Table I and plotted them with the red line for 100\% and the yellow line for 86.9\% effectiveness. The effective disinfection zone distribution in accordance with ranges is provided in Fig. \Ref{fig:luminocity_heatmap_plus_contour_plus_zones}.

Since UltraBot has a symmetrical design, the shape of the disinfection zone obtained from one side is also valid for the other side. Therefore, the final shape of a disinfection area resembles  butterfly.

\section{Coverage Path Planning Algorithm}
\subsection{The evolutionary approach}

To achieve an optimal trajectory, solving the online PPCR problem, we propose a GHACPP algorithm for the robotic disinfection by UV-C irradiation. %This approach is based on the Genetic Algorithms (GA) \cite{yakoubi2016path} which is inspired by evolution via natural selection, to solve optimization problems.
Our approach allows the disinfection robot to find the best part of the resulting disinfection trajectory, called mini-trajectory, obtained after each step of the target area exploration.

The global trajectory that allows the robot to cover the target area by the disinfection coverage spot fully consists of a set of mini-trajectories, obtained step by step along with the target area exploration and its digital map extension performed by Simultaneous Localization and Mapping (SLAM) \cite{hess2016real}. The mini-trajectories contain coordinates that are marked as visited after each mini-trajectory generation is completed. Each  mini-trajectory  is generated in a few steps according to Algorithm \ref{alg:evolutionary_approach}. 

% \vspace{-3em}

At the first step, our algorithm performs a generation of the initial population that consists of \textit{N} random mini-trajectories called chromosomes. Then, the chromosomes evolve by a set of mutations, applied with corresponding probabilities. After that, the mutated chromosomes are evaluated by a cost function and compared between each other and chromosomes of the previous population. \textit{N} chromosomes with the lowest costs form a new population, which can be mutated further. After the population is mutated \textit{M} times, a selection step is performed, where the chromosome with the lowest cost is selected as a new mini-trajectory that is added to the global robot trajectory. 

\begin{algorithm}[t]
\DontPrintSemicolon
\fontsize{9pt}{9.5pt}\selectfont
\caption{Evolutionary approach GHACPP of generating a mini-trajectory.}
\label{alg:evolutionary_approach}
\KwOutput{$MiniTrajectory$}
Generate an initial population $P_{0}(N)$. \\
$P\gets P_0$ \\
\For{$i \gets 1$ to $M$}    
    { 
    	$P_{mutated}\gets P$ \\
    	\For{$Mutation$ in $Mutations$}
        	{
        	    $P_{mutated}\gets Mutation(P_{mutated})$
        	}
        $P\gets P + P_{mutated}$ \\
        $Costs(P)\gets Evaluate(P)$ \\
        $P = Sort(P | Costs(P), N)$ \\
        \If{$P.first()$ is the same $T$ times in a row}
        {
            $MiniTrajectory\gets P.first()$ \\
            \Return {$MiniTrajectory$}
        }
    }
    $MiniTrajectory\gets P.first()$ \\
    \Return {$MiniTrajectory$}

\end{algorithm}

% \vspace{1em}

However, to reduce the computational costs spent on the mutation process, we propose to set the \textit{T} minimal amount of times when the best chromosome in terms of cost remains the same. If it occurs, the mutation process is stopped and followed by the selection step, where the chromosome is selected as a new mini-trajectory. As a result, there is always a chromosome that can be selected as a new part of the global robot trajectory to follow.

\subsection{Genetic mutations}
\label{sec:Genetic mutations}

Genetic mutations are operators that change some parts or even the whole input chromosome. As chromosomes are mini-trajectories, they consist of a set of points that are reached by the robot as it moves along a particular mini-trajectory. In this manner, when the chromosome is mutated, its point coordinates are modified. Crossover is another chromosome change operator in genetic algorithms that shuffles the chromosome parts of a population. However, the proposed algorithm implies the use of chromosomes with a small number of points (maximum 5), so the operation performed by crossover is covered by the proposed mutations. In this regard, we propose not to include it in the population mutation process, to save computational resources. A set of mutations used in the proposed UV-C Genetic-based Path Planner is described below.

\subsubsection{Random Sample Mutation} 
\begin{figure}[h]
\vspace{-0.5em}
\centering
\includegraphics[width=0.15\textwidth]{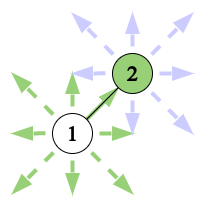}\label{true}
\vspace{-0.5em}
\caption{Allowed sampling directions while sampling the 2\textsuperscript{nd} point (green) and the 3\textsuperscript{rd} point (light blue).}
\label{fig:random_sample_mutation}
\vspace{-0.5em}
\end{figure}

This mutation completely replaces the input chromosome with a new one, leaving only the first point as it represents the current robot position. Having the first point, the second point is sampled in 8 possible directions with some resolution relative to the first point. Then, the sampling process is repeated with the second point,  excluding  the  direction  toward  the  first  point (see Fig. \ref{fig:random_sample_mutation}). After \textit{L} times of repeating the sampling process, an output chromosome consists of \textit{L + }1 points.

\subsubsection{Add Point Mutation} 
\begin{figure}[h]
\centering
\vspace{-0.5em}
\includegraphics[width=0.45\textwidth]{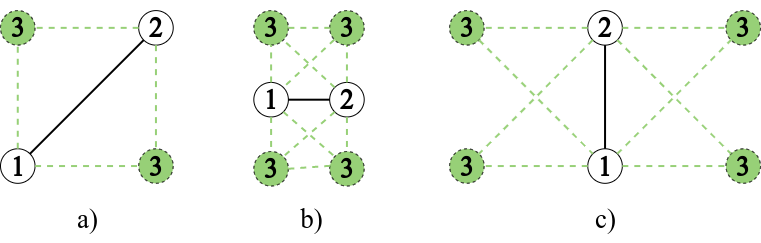}\label{true}
\vspace{-0.5em}
\caption{Allowed sampling positions (green) and connections (dash green) of the new added point, which is assigned to be intermediate between two chromosome points (1 and 2), located: a) diagonally, b) horizontally, c) vertically.}
\label{fig:add_point_mutation}
\vspace{-0.5em}
\end{figure}

This type of mutation allows adding one more point to the existing chromosome, increasing its length and repairing an unfeasible path. In contrast to \cite{hu2004knowledge} our approach suggests random selection of the chromosome point index where to add a new point. Thus, the new point can be either intermediate or the last one. Also, it is sampled with the same resolution as for Random Sample Mutation. If the point is intermediate, it should be sampled near both previous and next chromosome points to keep the resolution between chromosome points. All cases of adding a new intermediate point, are provided in Fig. \ref{fig:add_point_mutation}.

\subsubsection{Remove Point Mutation} 
\begin{figure}[h]
\vspace{-1em}
\centering
\includegraphics[width=0.45\textwidth]{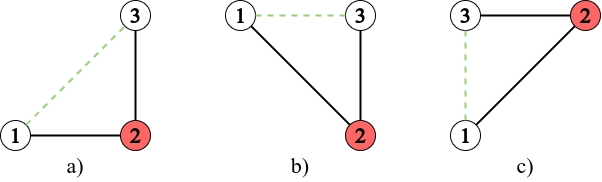}\label{true}
\caption{Allowed positions of the removed point (red) and new connections (dash green) between the previous and the next chromosome points (1 and 3), located: a) diagonally, b) horizontally, c) vertically.}
\vspace{-1em}
\label{fig:remove_point_mutation}
\end{figure}

This type of mutation works according to the principle opposite to Add Point Mutation. It removes a randomly selected point from the input chromosome, as in \cite{hu2004knowledge}. However, there are some restrictions. The first one is that the first point cannot be removed, as it represents the robot current position, and the robot is already in it. In addition, if the point remove index is not the last one in the chromosome, the point can be removed only if the previous and the next points can be connected with the same resolution as in previous two mutations. Situations, in which the point can be removed, are clarified in Fig. \ref{fig:remove_point_mutation}.

\subsection{Cost function}

A cost function is necessary for a detailed description of a generated chromosome $C$ (mini-trajectory). The function describes the absence of restrictions applicable to the trajectory, such as the occurrence of a collision, long trajectory length, the closeness of a robot to a human, etc. To evaluate the cost of each generated mini-trajectory, 6 penalties were taken into consideration. Thus, a sum of normalized and weighted penalties makes up the cost function (Algorithm \ref{alg:cost_function}).

Our algorithm is designed to take into account each penalty with a different weight. We assigned the highest weight to the \textit{Collision penalty} and \textit{Human closeness penalty}, because they are our top priority. The list of all penalties is given below in descending order according to their weights.

\textit{Collision penalty}. This penalty is responsible for the detection of collision that occurred while moving along the generated chromosome, Algorithm \ref{alg:cost_function}. A check of possible collisions is performed with some interpolation step, where the interpolated point is checked to be within an occupied or free area of a C-space. C-space is an occupancy map created by SLAM using LiDAR sensors, but with obstacles, which borders are extended on half of the robot size. It allows the collision detection algorithm not to recreate the robot shape in every trajectory point for a check but quickly perform the trajectory point check in the C-shape map instead. This approach allows reducing the required computational resources significantly. If the collision is detected, the collision penalty value is 1.0, otherwise, it equals 0.

\textit{Human closeness penalty} This penalty is responsible for determining whether the robot will bypass close to humans and irradiate them by UV-C lamps, following the current mini-trajectory. In order to estimate the robot closeness to humans, a Gaussian cost function is built for personal space \cite{fang2020human}, according to the following equation:

{
\begin{algorithm}[!t]
\fontsize{9pt}{9.5pt}\selectfont
\caption{Cost computation.}
\label{alg:cost_function}
\DontPrintSemicolon
  
  \KwInput{$MiniTrajectory$ (the chromosome)}
  \KwOutput{$Cost$}
  \KwData{$Step$, $W_1$ .. $W_6$ (weights), $DisinfectionMap$, $HumanPointCloud$}

$Cost\gets 0$ \\
$Interpolated = Interpolate(MiniTrajectory, Step)$ \\
\tcc{\textbf{Collision penalty:}} 
$Penalty_{1}\gets 0.0$ \\
\For{$Point$ in $Interpolated$}
{
    \If{$Point$ is in occupied area} {
        $Penalty_{1}\gets 1.0$ \\
    }
}

\tcc{\textbf{Human closeness penalty:}} 
$Penalty_{2}\gets 0.0$ \\
\For{$Point$ in $MiniTrajectory$} {
    \For{$HumanPoint$ in $HumanPointCloud$} 
    {   $PersonalCost\gets GetPersonalCost(Point, HumanPoint)$
        $Penalty_{2}\gets Max(Penalty_{2}, PersonalCost)$
    }

}
    
\tcc{\textbf{Visited penalty:}} 
$Penalty_{3}\gets 0.0$ \\
\For{$Point$ in $MiniTrajectory$} {
    \If{$Point$ has been already visited} {
        $Penalty_{3}\gets Penalty_{3} + 1.0$
    }
}
$Penalty_{3}\gets Normalize(Penalty_{3})$ \\

\tcc{\textbf{Absence of visited neighbour penalty:}} 
$Penalty_{4}\gets 0.0$ \\
\For{$Point$ in $MiniTrajectory$} {
    $Neighbours\gets GetNeighbourPoints(Point)$ \\
    \For{$Neighbour$ in $Neighbours$} {
        \If{$Neighbour$ has been already visited} {
            $Penalty_{4}\gets Penalty_{4} + 0.14$ \\
        }
    }
}
$Penalty_{4}\gets Normalize(Penalty_{4})$ \\

\tcc{\textbf{Turn penalty:}} 
$Penalty_{5}\gets 0.0$ \\
\For{$i \gets 2$ to $MiniTrajectory.size()$}
{
    $Turn\gets MiniTrajectory(i).angle - MiniTrajectory(i-1).angle$ \\
    \If{$Turn = 0^{\circ}$} {
        $Penalty_{5}\gets Penalty_{5} + 0.0$ \\
    }
    {0$^{\circ} <$ Turn $\leq$ 45$^{\circ}$}
    {
        $Penalty_{5}\gets Penalty_{5} + 1.0$ \\
    }
    {45$^{\circ} <$ Turn $\leq$ 90$^{\circ}$} {
        $Penalty_{5}\gets Penalty_{5} + 0.5$ \\
    }
    {90$^{\circ} <$ Turn $\leq$ 180$^{\circ}$} {
        $Penalty_{5}\gets Penalty_{5} + 1.0$ \\
    }
}
$Penalty_{5} \gets Normalize(Penalty_{5})$ \\

\tcc{\textbf{Repeating disinfection penalty:}} 
$Penalty_{6}\gets 0.0$ \\
\For{$Point$ in $Interpolated$}
{
    $Mask\gets GetDisinfectionMask(Point)$ \\
    $PointPenalty\gets DisinfectionMap.apply(Mask)$ \\
    $Penalty_{6} \gets Penalty_{6} + PointPenalty$ \\
}
$Penalty_{6}\gets Normalize(Penalty_{6})$ \\
\tcc{\textbf{Cost function:}} 
$Cost\gets \sum_{i=1}^{6} W_{i} * Penalty_{i}$ \\
\Return {$Cost$} \\
\end{algorithm}
}

% \vspace{-1.2 em}
\vspace{-1 em}
\begin{equation}
\centering
\begin{multlined}
PersonalCost_{i}(Point(x_{i}, y_{i})) = e{^{-\dfrac{d^2}{2\sigma_{i}}}},
\end{multlined}
\label{eq:personal_cost}
\end{equation}

\noindent where $d = \sqrt{(x - x_i)^{2} + (y - y_i)^{2}}$ is the distance to a person, $x_{i}$, $y_{i}$ are the coordinates of the person $p_{i}$ location, and $\sigma_{i}$ is the standard deviation of the Gaussian cost function, which prohibits the robot from crossing the personal space and satisfies the requirement of safety. The cost function is applied to each point of the Human Point Cloud obtained from RGB-D cameras. The points with the highest costs affect the planning of the robot's trajectory and force the robot to be attentive to humans.

\begin{figure}[!t]
\centering
\vspace{0.5em}
\includegraphics[width=0.4\textwidth]{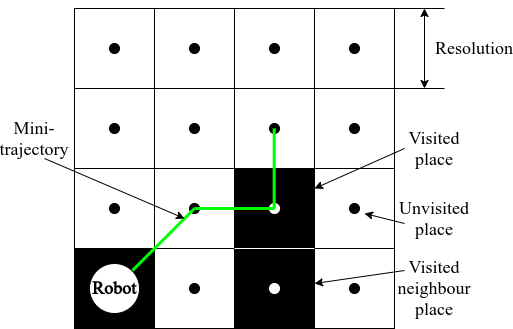}
\caption{The map of the visited places (mini-trajectory points).}
\label{fig:visited_map}
\vspace{-1.5 em}
\end{figure}

\textit{Visited penalty}. This penalty is responsible for determining whether the robot will revisit places it has already visited, following previously generated mini-trajectories that are part of a global trajectory already traveled. As the places were visited with some resolution, proposed in \autoref{sec:Genetic mutations}, the fact of revisiting is identified with the use of a cost map of visited places, provided in Fig. \ref{fig:visited_map}. If already visited points, except the first one, are in the newly generated chromosome, the normalized penalty is applied (line 11 of Algorithm \ref{alg:cost_function}).
     
\textit{Absence of visited neighbour penalty}. This penalty is responsible for covering the target area that remained uncleaned after the robot pass. These uncleaned areas appear because of UV-C lamps installed on the left and right sides of the robot only, which leads to not disinfecting areas, along which the robot actually moves. Performing boustrophedon-like motion allows the robot to disinfect the local area fully. This motion is achieved by applying the penalty in cases when there are no visited neighbor places for the newly generated mini-trajectory point (Fig. \ref{fig:visited_map}).
   
\textit{Turn penalty}. This parameter is responsible for reducing the number of turns performed while performing the disinfection task and following the generated mini-trajectories. Whereas the turn penalty value for the whole chromosome (mini-trajectory) remains normalized, absolute penalty values for each point are distributed according to rules in line 25 of Algorithm \ref{alg:cost_function}. This distribution allows to generate mini-trajectories, that are as straight as it is possible.

\textit{Repeating disinfection penalty}. This penalty is responsible for detection of disinfection of already disinfected areas. 
While the target area is covered with some robot disinfection zone shape, the corresponding map cells are marked as already disinfected. The repeating disinfection penalty is computed by drawing the disinfection zone in each interpolated point of the mini-trajectory with the further comparison of the obtained disinfection masks with the current disinfection map.

\section{Experiments}

\subsection{Experiment 1}
\subsubsection{Experimental setup}

\begin{figure}[h]
\centering
\vspace{-1em}
\includegraphics[width=0.37\textwidth]{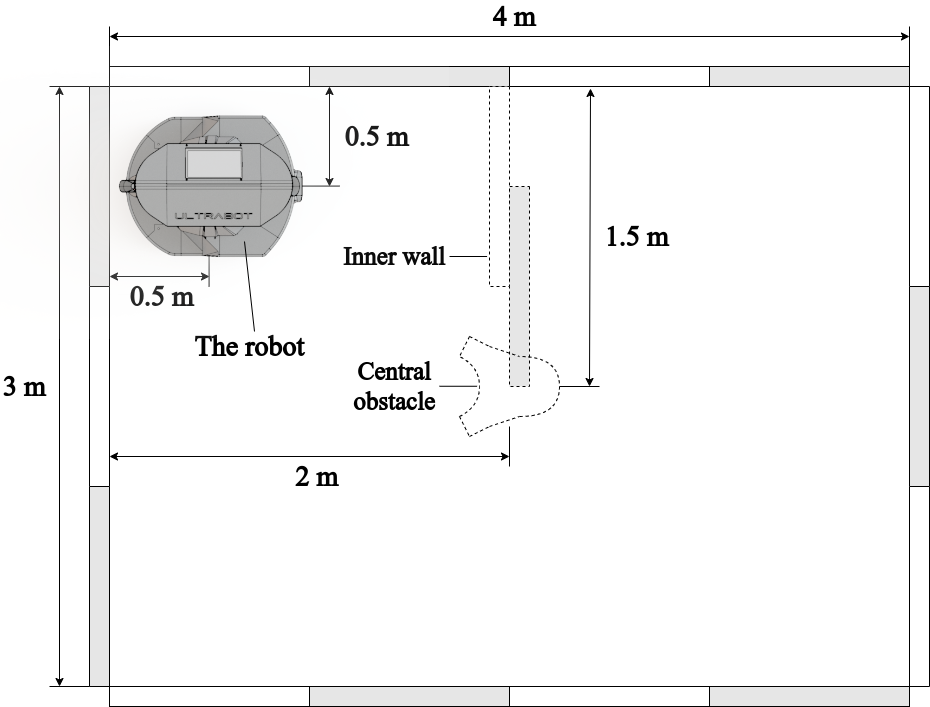}\label{true}
\caption{Experimental area for 3 scenarios: 1) Central obstacle presence. 2) Inner wall presence. 3) The area is empty.}
\vspace{-1em}
\label{fig:experimental_area}
\end{figure}

In order to validate the performance of the developed GHACPP algorithm, we set up an experimental area measuring 3 by 4 meters and prepared three experimental scenarios  according to Fig. \ref{fig:experimental_area}.
\begin{comment}
\begin{itemize}
    \item Scenario 1 - presence of the obstacle in the area center;
    \item Scenario 2 - presence of the inner wall;
    \item Scenario 3 - absence of any obstacles.
\end{itemize}
\end{comment}

\begin{figure*} [!t]
\vspace{0.5em}
\begin{center}
\includegraphics[width=0.9\textwidth]{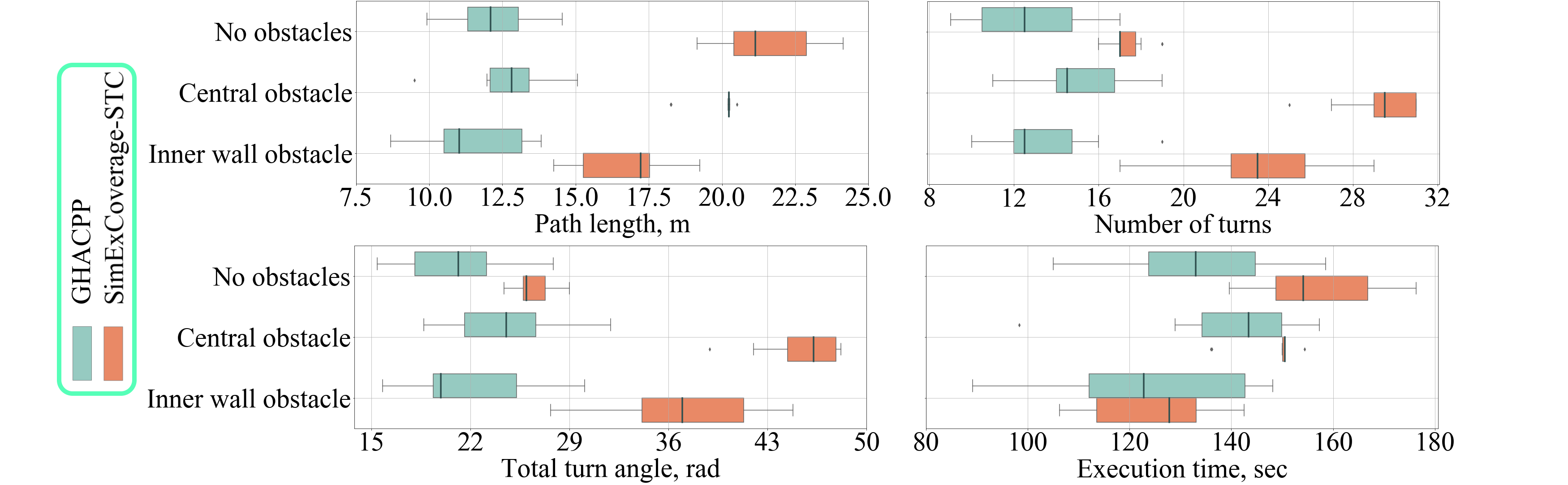}
%\vspace{-0.5em}
\caption{Results of the first experiment for path planning evaluation.}
\label{fig:all}
\vspace{-2em}
\end{center}
\end{figure*}

Performance of the GHACPP algorithm, solving an online PPCR problem for a mobile robot, was compared (Table \ref{tab:area_coverage} and Fig. \ref{fig:all}) with the state-of-the-art algorithm called SimExCoverage-STC \cite{falaki2020simultaneous}. SimExCoverage-STC is one of the most recent online coverage path planning algorithms, which combines solving both exploration and coverage problems. The exploration provides a map for coverage path planning, while the coverage provides the path for robot motion. While the robot is moving along the coverage path, exploration is carried out.

The UltraBot was launched from the left upper corner 10 times under both GHACPP and SimExCoverage-STC algorithms for each scenario. The robot's speed was set up to 0.5 km/h. The following parameters, reflecting the performance of the path planner, were calculated and compared:
\begin{enumerate}
    \item Area coverage, \%;
    \item Path length, m;
    \item Number of turns;
    \item Total turn angle, rad;
    \item Time, sec.
\end{enumerate}

To compute the \textit{Repetition disinfection penalty} and the disinfected area percentage, a special disinfection zone shape of the robot was used. In the experiment, a zone that provides 84.6\% of disinfection effectiveness was considered (Fig. \ref{fig:luminocity_heatmap_plus_contour_plus_zones}).

\subsubsection{Scenario 1}
The first experimental scenario was conducted with an obstacle placed in the center of the experimental area.
The GHACCP algorithm disinfected 0.3\% less area, whereas it significantly reduced the overall trajectory length by 36.0\%, the number of turns by 48.8\%, the total turn angle by 45.2\%, and disinfection mission time by 5.9\%.

\subsubsection{Scenario 2}
 
The second experimental scenario was conducted with the inner wall, which doesn't allow the robot SLAM algorithms to reconstruct a much larger part of the experimental area in the beginning, in comparison with the first scenario. 
The experimental results showed that with 0.6\% less area disinfected, the GHACPP algorithm significantly reduced the overall trajectory length by 31.5\%, the number of turns by 43.0\%, the total turn angle by 39.9\%, and disinfection mission time by 0.7\%.

\subsubsection{Scenario 3}

The third experiment allowed to estimate performance of both planners in the static and totally known environment. 
The experimental results showed that with 0.9\% less area disinfected, the GHACPP algorithm significantly reduced the overall trajectory length by 43.8\%, the number of turns by 26.6\%, the total turn angle by 20.6\%, and disinfection mission time by 16.1\%.

\subsubsection{Result of the 1st Experiment}
It can be concluded that, in total, with less than 1\% area coverage (0.6\% in average), the GHACPP algorithm allowed to reduce the total path length by 43.8\% (37.1\% in average), the number of turns by 48.8\% (39.5\% in average), the total turn angle by 45.2\% (35.2\% in average), and coverage time by 16.1\% (7.6\% in average), in comparison with the state-of-the-art SimExCoverage-STC algorithm (Table \ref{tab:area_coverage}, Fig. \ref{fig:all}). 

The GHACPP algorithm achieved the lowest performance in total path length and execution time with the inner wall presence. More exploration map update steps were required due to the large obstacle dimensions. In an empty area, the highest performance for these parameters was achieved, as the SimExCoverage-STC algorithm made many unnecessary bypasses that led to high disinfection repeatability of already disinfected areas. 

\begin{table}[h]
%\vspace{-1em}
    \centering
    \caption{\footnotesize\scshape Coverage Area.}
    \begin{tabular}{|c|c|c|c|c|}
    \hline
    Scenario № & Algorithm & Mean & Min & SD\\
    \hline
    \multirow{2}{*}{1} & GHACPP & 99.60 & 98.43 & 0.48 \\
    \cline{2-5}
     & SimExCoverage-STC & 99.98 & 99.84 & 0.05 \\
     
    \hline
    \multirow{2}{*}{2} & GHACPP & 99.20 & 97.91 & 0.82 \\
    \cline{2-5}
     & SimExCoverage-STC & 99.77 & 98.89 & 0.42 \\
     
    \hline
    \multirow{2}{*}{3} & GHACPP & 99.09 & 97.67 & 0.81 \\
    \cline{2-5}
     & SimExCoverage-STC & 100.00 & 99.98 & 0.01 \\
    \hline
    \end{tabular}
\label{tab:area_coverage}
\vspace{-1.5em}    
\end{table}

\subsection{Experiment 2}
The first experiment was conducted in the small size environment to validate the algorithm's performance statistically. However, no human presence was considered in this case. Being in the robot area of view, humans must affect the disinfection process, as the robot must provide safety for humans and not irradiate them with UV-C lamps. In order to consider both the larger environment and human presence in the disinfection task, 2 experimental scenarios were prepared for an area measuring 6 by 14.5 meters. 

The UltraBot was launched from the left lower corner 1 time for each of two path-planning algorithms for each scenario. The robot's speed was set up to 0.5 km/h. Additionally, as for the first experiment, we set the same robot disinfection zone shape and computed the same parameters, representing path planner performance.

\subsubsection{Scenario 1}
\begin{figure}[!h]

\centering
\begin{subfigure}{0.9\linewidth}
% \vspace{-0.5em}
\centering
\includegraphics[width=\textwidth]{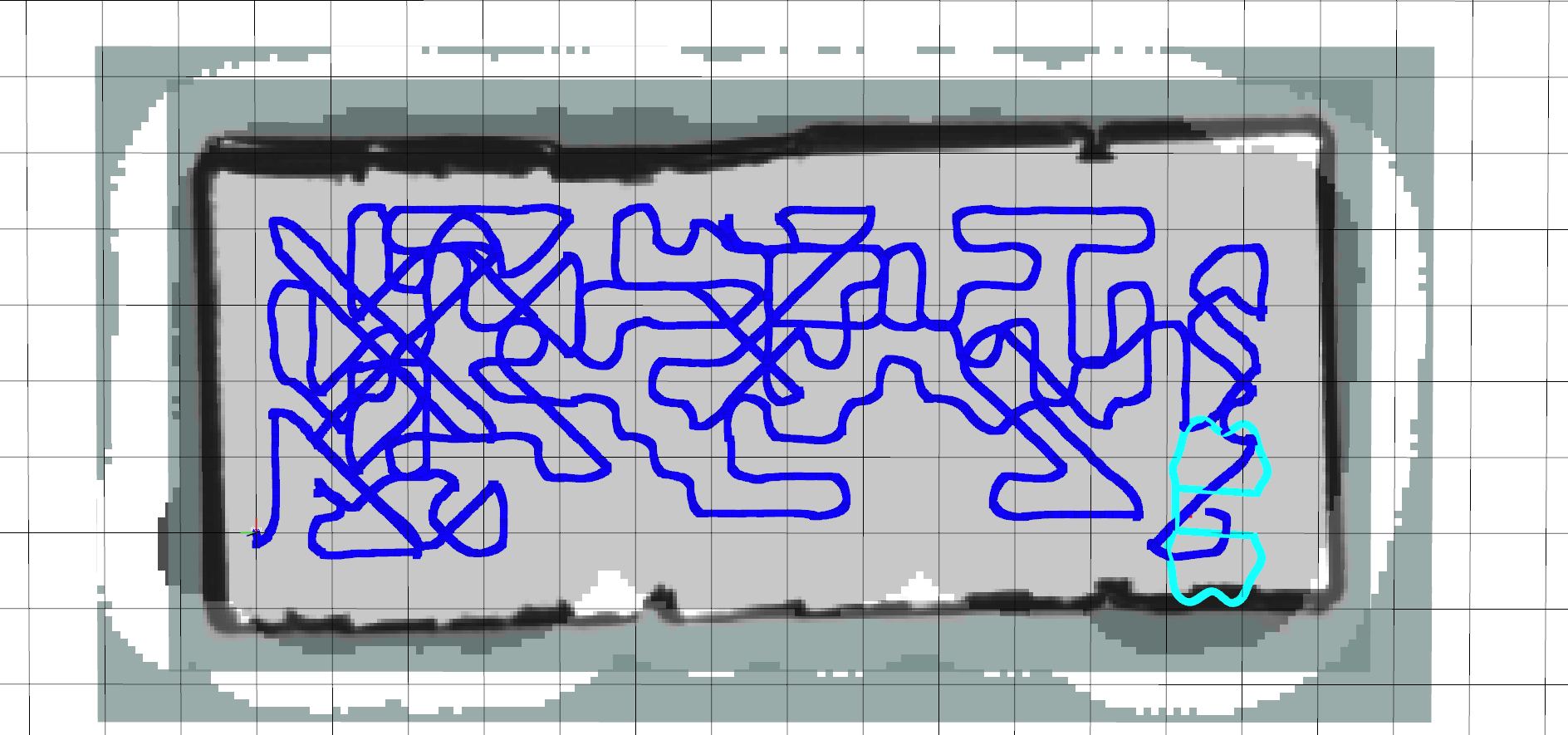}
\caption{GHACPP: Area coverage = 98.9\%, Path length = 188.97 m.}
\end{subfigure}
\begin{subfigure}{0.9\linewidth}
\centering
\includegraphics[width=\textwidth]{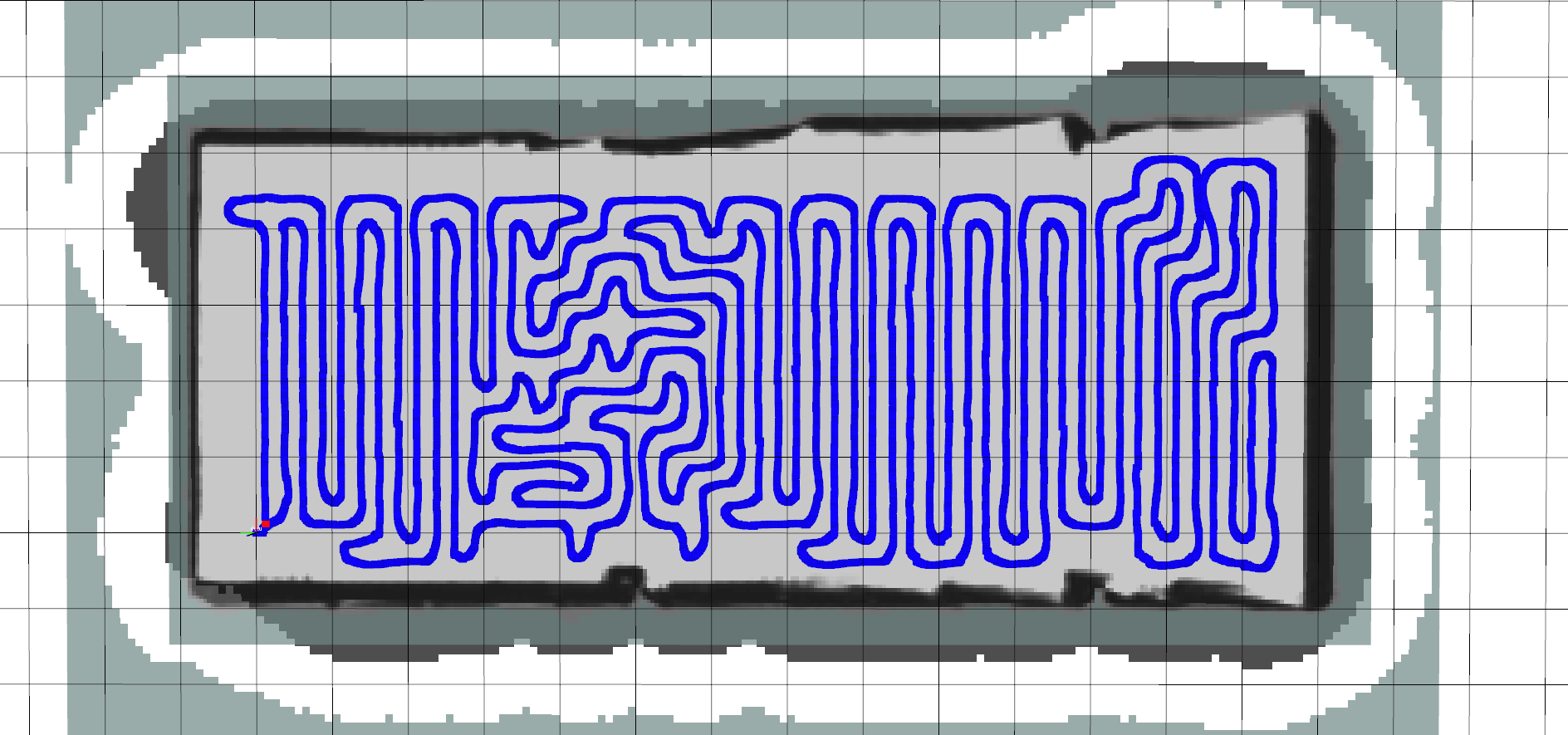}
\caption{SimExCoverage-STC: Area coverage = 100.0\%, Path length = 266.32 m.}
\end{subfigure}
\caption{Algorithm trajectories (blue lines) in the empty area. The dark shaded area is the disinfected area covered by the robot disinfection zone (light blue counter in the right bottom corner of Fig. 9(a).}
\label{fig:exp2_trajectories_empty}
\vspace{-2 em}
\end{figure}

The first experimental scenario allowed to estimate generated trajectories and disinfection results for both planners in a large static environment without humans. Robot trajectories are provided in Fig. \ref{fig:exp2_trajectories_empty}. As a result, with 1.1\% less area disinfected the GHACPP algorithm reduced the overall trajectory length by 29.0\%, a number of turns by 14.0\%, and a total turn angle by 8.5\%, in comparison with the SimExCoverage-STC algorithm.

\subsubsection{Scenario 2}
The second experimental scenario was conducted considering 3 persons in protective clothing located in the center and near two opposite walls of the experimental area. This scenario allowed to estimate planner performance in human safety while performing a disinfection task. Corresponding robot trajectories are provided in Fig. \ref{fig:exp2_trajectories_human}. 
According to the experimental results, with 19.7\% less disinfected area, the GHACPP algorithm showed a lower total trajectory length by 62.5\%, a lower number of turns by 54.8\%, and a lower total turn angle 54.9\% and lower execution time by 50.3\%. The algorithm showed less performance in the covered area percentage, as it didn't allow the robot to irradiate people during disinfection, as opposed to the SimExCoverage-STC algorithm, which completely irradiate people several times in one disinfection task.

\subsubsection{Result of the 2nd Experiment}
Working in a large space without humans, the GHACPP algorithm showed the lower total path length, the number of turns, and the total turn angle compared to the SimExCoverage-STC algorithm. Performing disinfection in the large space side by side with humans, both path planning algorithms successfully avoided collisions with them. However, it should be mentioned, that the GHACPP algorithm did not harm people by UV-C irradiation, as it considered human presence in the area. In contrast, the SimExCoverage-STC algorithm fully irradiated all humans in the area. Therefore, we can conclude that the GHACPP algorithm is able to perform safe disinfection effectively, working in the same space as humans, while not irradiating them by UV-C light, in contrast to the compared path planning algorithm.

\section{Conclusion}

In this work, a Genetic-based Human-Aware Coverage Path Planning algorithm was developed for a mobile differential-drive robot with a non-regular UV-C lamp arrangement to disinfect premises, working in the same space as humans. This algorithm was implemented for the high-level control system of the UltraBot, an autonomous indoor mobile disinfection robot with two independent blocks of UV-C lamps. An experiment with three different scenarios on the path planner performance validation was conducted in a limited area with obstacles of different sizes. Additionally, an experiment on the planner performance in a large space, considering human presence, was held. 
The experimental results confirmed both the high level of safety for humans and the efficiency of the developed algorithm in terms of path length (37.1\%), number (39.5\%) and size (35.2\%) of turns, and time (7.6\%) to complete the disinfection task, with a small loss in the percentage of area covered (0.6\%), in comparison with one of the latest state-of-the-art approach.

In the future, we plan to conduct several ablation studies to expand the scope of this algorithm. Our method can be applied in many areas of robotics, where it is necessary to solve the problem of coverage path planning for the completeness of tasks with the presence of any obstacles, for example, research and search for objects using drones \cite{yatskin2017principles}, robots for detecting plant diseases \cite{karpyshev2021autonomous}, any kind of inventory in warehouses \cite{kalinov2019high, kalinov2020warevision} and stores \cite{petrovsky2020customer}, and even for the Mars exploration rovers \cite{petrovsky2022two}.

\bibliographystyle{IEEEtran}
\bibliography{ppcr-refs}

\begin{thebibliography}{10}
\providecommand{\url}[1]{#1}
\csname url@samestyle\endcsname
\providecommand{\newblock}{\relax}
\providecommand{\bibinfo}[2]{#2}
\providecommand{\BIBentrySTDinterwordspacing}{\spaceskip=0pt\relax}
\providecommand{\BIBentryALTinterwordstretchfactor}{4}
\providecommand{\BIBentryALTinterwordspacing}{\spaceskip=\fontdimen2\font plus
\BIBentryALTinterwordstretchfactor\fontdimen3\font minus
  \fontdimen4\font\relax}
\providecommand{\BIBforeignlanguage}[2]{{%
\expandafter\ifx\csname l@#1\endcsname\relax
\typeout{** WARNING: IEEEtran.bst: No hyphenation pattern has been}%
\typeout{** loaded for the language `#1'. Using the pattern for}%
\typeout{** the default language instead.}%
\else
\language=\csname l@#1\endcsname
\fi
#2}}
\providecommand{\BIBdecl}{\relax}
\BIBdecl

\bibitem{tiseni2021uv}
L.~Tiseni, D.~Chiaradia, M.~Gabardi, M.~Solazzi, D.~Leonardis, and A.~Frisoli,
  ``Uv-c mobile robots with optimized path planning: Algorithm design and
  on-field measurements to improve surface disinfection against sars-cov-2,''
  \emph{IEEE Robotics \& Automation Magazine}, vol.~28, no.~1, pp. 59--70,
  2021.

\bibitem{chanprakon2019ultra}
P.~Chanprakon, T.~Sae-Oung, T.~Treebupachatsakul, P.~Hannanta-Anan, and
  W.~Piyawattanametha, ``An ultra-violet sterilization robot for
  disinfection,'' in \emph{2019 5th International Conference on Engineering,
  Applied Sciences and Technology (ICEAST)}.\hskip 1em plus 0.5em minus
  0.4em\relax IEEE, 2019, pp. 1--4.

\bibitem{UVDrobots}
\BIBentryALTinterwordspacing
``Enhanced cleaning routines. {UVD Robots},'' 2021. [Online]. Available:
  \url{https://www.uvd-robots.com/}
\BIBentrySTDinterwordspacing

\bibitem{ultrabot}
S.~Perminov, N.~Mikhailovskiy, A.~Sedunin, I.~Okunevich, I.~Kalinov,
  M.~Kurenkov, and D.~Tsetserukou, ``Ultrabot: {A}utonomous {M}obile {R}obot
  for {I}ndoor {UV-C} {D}isinfection,'' in \emph{2021 17th International
  Conference on Automation Science and Engineering (CASE)}.\hskip 1em plus
  0.5em minus 0.4em\relax IEEE, 2021.

\bibitem{luo2020research}
Q.~Luo, H.~Wang, Y.~Zheng, and J.~He, ``Research on path planning of mobile
  robot based on improved ant colony algorithm,'' \emph{Neural Computing and
  Applications}, vol.~32, no.~6, pp. 1555--1566, 2020.

\bibitem{liu2015robot}
J.~Liu, J.~Yang, H.~Liu, P.~Geng, and M.~Gao, ``Robot global path planning
  based on ant colony optimization with artificial potential field,''
  \emph{Transactions of the Chinese Society for Agricultural Machinery},
  vol.~46, no.~9, pp. 18--27, 2015.

\bibitem{wang2020neural}
J.~Wang, W.~Chi, C.~Li, C.~Wang, and M.~Q.-H. Meng, ``Neural rrt*:
  Learning-based optimal path planning,'' \emph{IEEE Transactions on Automation
  Science and Engineering}, vol.~17, no.~4, pp. 1748--1758.

\bibitem{wang2020optimal}
J.~Wang and M.~Q.-H. Meng, ``Optimal path planning using generalized voronoi
  graph and multiple potential functions,'' \emph{IEEE Transactions on
  Industrial Electronics}, vol.~67, no.~12, pp. 10\,621--10\,630, 2020.

\bibitem{gabriely2001spanning}
Y.~Gabriely and E.~Rimon, ``Spanning-tree based coverage of continuous areas by
  a mobile robot,'' \emph{Annals of mathematics and artificial intelligence},
  vol.~31, no.~1, pp. 77--98, 2001.

\bibitem{gonzalez2005bsa}
E.~Gonzalez, O.~Alvarez, Y.~Diaz, C.~Parra, and C.~Bustacara, ``Bsa: A complete
  coverage algorithm,'' in \emph{Proceedings of the IEEE International
  Conference on Robotics and Automation}.\hskip 1em plus 0.5em minus
  0.4em\relax IEEE, 2005, pp. 2040--2044.

\bibitem{galceran2013survey}
E.~Galceran and M.~Carreras, ``A survey on coverage path planning for
  robotics,'' \emph{RAS}, vol.~61, no.~12, pp. 1258--1276, 2013.

\bibitem{bormann2018indoor}
R.~Bormann, F.~Jordan, J.~Hampp, and M.~H{\"a}gele, ``Indoor coverage path
  planning: Survey, implementation, analysis,'' in \emph{2018 IEEE
  International Conference on Robotics and Automation (ICRA)}.\hskip 1em plus
  0.5em minus 0.4em\relax IEEE, 2018, pp. 1718--1725.

\bibitem{cai2014research}
Z.~Cai, S.~Li, Y.~Gan, R.~Zhang, and Q.~Zhang, ``Research on complete coverage
  path planning algorithms based on a* algorithms,'' \emph{The Open Cybernetics
  \& Systemics Journal}, vol.~8, no.~1, 2014.

\bibitem{song2018varepsilon}
J.~Song and S.~Gupta, ``e: An online coverage path planning algorithm,''
  \emph{IEEE Transactions on Robotics}, vol.~34, no.~2, pp. 526--533, 2018.

\bibitem{yakoubi2016path}
M.~A. Yakoubi and M.~T. Laskri, ``The path planning of cleaner robot for
  coverage region using genetic algorithms,'' \emph{Journal of innovation in
  digital ecosystems}, vol.~3, no.~1, pp. 37--43, 2016.

\bibitem{prayash2019designing}
H.~S.~H. Prayash, M.~R. Shaharear, M.~F. Islam, S.~Islam, N.~Hossain, and
  S.~Datta, ``Designing and optimization of an autonomous vacuum floor cleaning
  robot,'' in \emph{2019 IEEE RAAICON}.\hskip 1em plus 0.5em minus 0.4em\relax
  IEEE, pp. 25--30.

\bibitem{vega2019socially}
A.~Vega, L.~J. Manso, D.~G. Macharet, P.~Bustos, and P.~N{\'u}{\~n}ez,
  ``Socially aware robot navigation system in human-populated and interactive
  environments based on an adaptive spatial density function and space
  affordances,'' \emph{Pattern Recognition Letters}, vol. 118.

\bibitem{cai2019adaptive}
K.~Cai, C.~Wang, C.~Li, S.~Song, and M.~Q.-H. Meng, ``Adaptive sampling for
  human-aware path planning in dynamic environments,'' in \emph{2019 IEEE
  International Conference on Robotics and Biomimetics (ROBIO)}.\hskip 1em plus
  0.5em minus 0.4em\relax IEEE, 2019, pp. 1987--1994.

\bibitem{perez2018learning}
N.~P{\'e}rez-Higueras, F.~Caballero, and L.~Merino, ``Learning human-aware path
  planning with fully convolutional networks,'' in \emph{2018 IEEE ICRA}.\hskip
  1em plus 0.5em minus 0.4em\relax IEEE, 2018, pp. 5897--5902.

\bibitem{fang2020human}
F.~Fang, M.~Shi, K.~Qian, B.~Zhou, and Y.~Gan, ``A human-aware navigation
  method for social robot based on multi-layer cost map,'' \emph{International
  Journal of Intelligent Robotics and Applications}, vol.~4, no.~3, pp.
  308--318, 2020.

\bibitem{falaki2020simultaneous}
P.~M.~M. Falaki, A.~Padman, V.~G. Nair, and K.~Guruprasad, ``Simultaneous
  exploration and coverage by a mobile robot,'' in \emph{Control
  Instrumentation Systems}.\hskip 1em plus 0.5em minus 0.4em\relax Springer,
  2020, pp. 33--41.

\bibitem{buonanno2020far}
M.~Buonanno, D.~Welch, I.~Shuryak, and D.~J. Brenner, ``Far-uvc light (222 nm)
  efficiently and safely inactivates airborne human coronaviruses,''
  \emph{Scientific Reports}, vol.~10, no.~1, pp. 1--8, 2020.

\bibitem{bv2006ultraviolet}
P.~L. BV, ``Ultraviolet purification application information,'' 2006.

\bibitem{ultrabot_zone}
N.~Mikhailovskiy, A.~Sedunin, S.~Perminov, I.~Kalinov, and D.~Tsetserukou,
  ``Ultrabot: {A}utonomous {M}obile {R}obot for {I}ndoor {UV-C}
  {D}isinfection,'' in \emph{2021 26th International Conference on Emerging
  Technologies and Factory Automation (ETFA)}.\hskip 1em plus 0.5em minus
  0.4em\relax IEEE, 2021.

\bibitem{hess2016real}
W.~Hess, D.~Kohler, H.~Rapp, and D.~Andor, ``Real-time loop closure in 2d lidar
  slam,'' in \emph{IEEE ICRA}.\hskip 1em plus 0.5em minus 0.4em\relax IEEE,
  2016, pp. 1271--1278.

\bibitem{hu2004knowledge}
Y.~Hu and S.~X. Yang, ``A knowledge based genetic algorithm for path planning
  of a mobile robot,'' in \emph{IEEE ICRA Proceedings 2004}, vol.~5.\hskip 1em
  plus 0.5em minus 0.4em\relax IEEE, 2004, pp. 4350--4355.

\bibitem{yatskin2017principles}
D.~Yatskin and I.~Kalinov, ``Principles of solving the space monitoring problem
  by multirotors swarm,'' in \emph{2017 IVth International Conference on
  Engineering and Telecommunication (EnT)}.\hskip 1em plus 0.5em minus
  0.4em\relax IEEE, 2017, pp. 47--50.

\bibitem{karpyshev2021autonomous}
P.~Karpyshev, V.~Ilin, I.~Kalinov, A.~Petrovsky, and D.~Tsetserukou,
  ``Autonomous mobile robot for apple plant disease detection based on cnn and
  multi-spectral vision system,'' in \emph{2021 IEEE/SICE international
  symposium on system integration (SII)}.\hskip 1em plus 0.5em minus
  0.4em\relax IEEE, 2021, pp. 157--162.

\bibitem{kalinov2019high}
I.~Kalinov, E.~Safronov, R.~Agishev, M.~Kurenkov, and D.~Tsetserukou,
  ``High-precision uav localization system for landing on a mobile
  collaborative robot based on an ir marker pattern recognition,'' in
  \emph{2019 IEEE 89th Vehicular Technology Conference (VTC2019-Spring)}.\hskip
  1em plus 0.5em minus 0.4em\relax IEEE, 2019, pp. 1--6.

\bibitem{kalinov2020warevision}
I.~Kalinov, A.~Petrovsky, V.~Ilin, E.~Pristanskiy, M.~Kurenkov, V.~Ramzhaev,
  I.~Idrisov, and D.~Tsetserukou, ``Warevision: Cnn barcode detection-based uav
  trajectory optimization for autonomous warehouse stocktaking,'' \emph{IEEE
  Robotics and Automation Letters}, vol.~5, no.~4, pp. 6647--6653, 2020.

\bibitem{petrovsky2020customer}
A.~Petrovsky, I.~Kalinov, P.~Karpyshev, M.~Kurenkov, V.~Ramzhaev, V.~Ilin, and
  D.~Tsetserukou, ``Customer behavior analytics using an autonomous
  robotics-based system,'' in \emph{2020 16th International Conference on
  Control, Automation, Robotics and Vision (ICARCV)}.\hskip 1em plus 0.5em
  minus 0.4em\relax IEEE, 2020, pp. 327--332.

\bibitem{petrovsky2022two}
A.~Petrovsky, I.~Kalinov, P.~Karpyshev, D.~Tsetserukou, A.~Ivanov, and
  A.~Golkar, ``The two-wheeled robotic swarm concept for mars exploration,''
  \emph{Acta Astronautica}, vol. 194, pp. 1--8, 2022.

\end{thebibliography}

\end{document}